\documentclass[final,5p,times,singlecolumn]{elsarticle}
\usepackage{amssymb}
\usepackage{amsmath}
\usepackage{amsfonts}
\usepackage{picins}
\usepackage{subfigure}
\usepackage{graphicx}
\usepackage{amsfonts}
\usepackage{amsmath}
\usepackage{xcolor}
\usepackage{booktabs}
\usepackage{graphicx}
\usepackage{caption}
\usepackage{ulem}
\usepackage{soul}
\graphicspath{{image/}}
\DeclareGraphicsExtensions{.pdf,.jpeg,.png,eps}
\usepackage{multirow}
\usepackage{color}
\usepackage{multicol}

\usepackage[switch]{lineno}

\journal{NEUROCOMPUTING}

\begin{document}
	
	\setstcolor{red}
	\title{A Joint Model for IT Operation Series Prediction and Anomaly Detection}
	\author[mymainaddress]{Run-Qing Chen}
	\author[mysecondaddress]{Guang-Hui Shi}
	\author[mymainaddress]{Wan-Lei~Zhao\corref{mycorrespondingauthor}}
	\cortext[mycorrespondingauthor]{Corresponding author: Wan-Lei Zhao}
	\ead{wlzhao@xmu.edu.cn}
	\author[mymainaddress]{Chang-Hui~Liang}
	
	\address[mymainaddress]{Fujian Key Laboratory of Sensing and Computing for Smart City, \\ School of Information Science and Engineering, Xiamen University, \\Xiamen 361005, Fujian, China}
	\address[mysecondaddress]{Bonree Inc., Beijing, China}
	
	\begin{abstract}
		Status prediction and anomaly detection are two fundamental tasks in automatic IT systems monitoring. In this paper, a joint model Predictor \& Anomaly Detector (PAD) is proposed to address these two issues under one framework. In our design, the variational auto-encoder (VAE) and long short-term memory (LSTM) are joined together. The prediction block (LSTM) takes clean input from the reconstructed time series by VAE, which makes it robust to the anomalies and noise for prediction task. In the meantime, the LSTM block maintains the long-term sequential patterns, which are out of the sight of a VAE encoding window. This leads to the better performance of VAE in anomaly detection than it is trained alone. In the whole processing pipeline, the spectral residual analysis is integrated with VAE and LSTM to boost the performance of both. The superior performance on two tasks is confirmed with the experiments on two challenging evaluation benchmarks.
	\end{abstract}
	
	
	\maketitle

	\section{Introduction}
	Due to the steady growth of cloud computing and the wide-spread of various web services, a big volume of IT operation data are generated daily. IT operations analytics aims to discover patterns from these huge amounts of time series data. Such that it is able to automate or monitor an IT system based on the operation data. It is widely known as artificial intelligence for IT operations (AIOps)~\cite{Dang2019}. It has been explored in recent works~\cite{Chen2019,Laptev2017,Ren2019,Siffer2017,Xu2018,Zhu2017}. Two fundamental tasks in AIOps are future status prediction and anomaly detection on the key performance indicators (KPIs), such as the time series about the number of user accesses and memory usage, etc.
	
	\begin{figure*}[!t]
		\centering
		\subfigure[hour-level \& stable]{\includegraphics[width=0.25\linewidth]{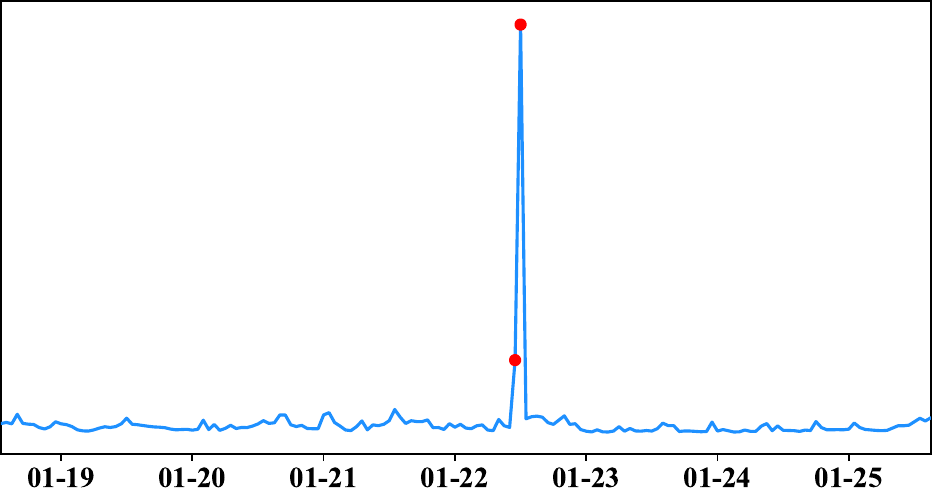}}
		\hspace{0.08in}
		\subfigure[hour-level \& periodic]{\includegraphics[width=0.25\linewidth]{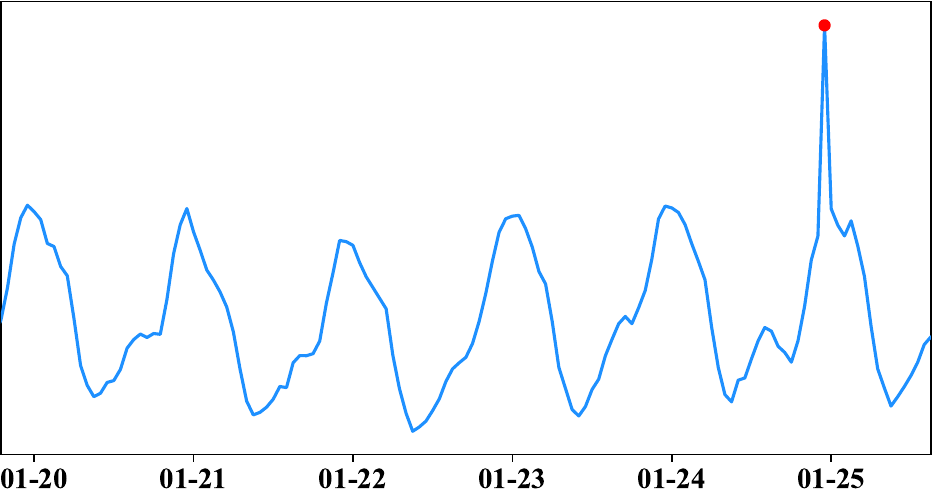}}\\
		\subfigure[minute-level \& stable]{\includegraphics[width=0.25\linewidth]{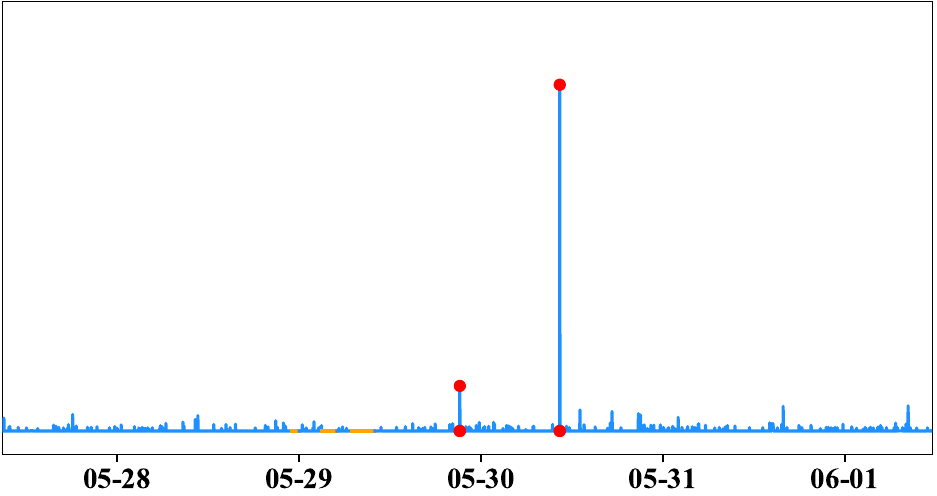}}
		\hspace{0.08in}
		\subfigure[minute-level \& periodic]{\includegraphics[width=0.25\linewidth]{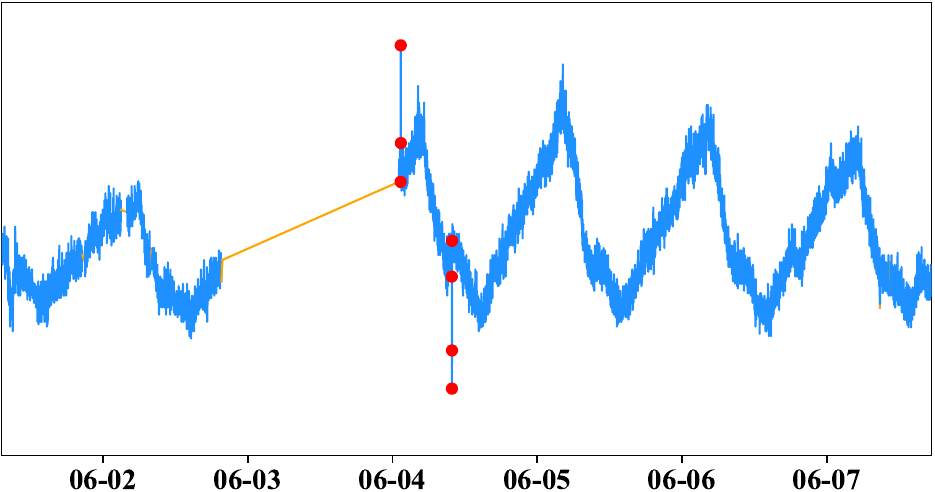}}
		\centering
		\caption{Sample segments from four KPI series. The anomalies are shown in red and missing statuses are shown in orange.}
		\label{fig:1}
	\end{figure*}
	
	In general, a sequence of KPIs is given as a univariate time series $X=\{x_1, \cdots, x_t, x_{t+1}, \cdots, x_{n-1}, x_n\}$, where the subscript represents the timestamp and $x_t$ is the real-valued status at one timestamp. Given the statuses from timestamp \textit{1} to $t$ are known, the anomaly detection is to judge whether the status on timestamp $t$ significantly differs from the majority of a time series. While the prediction is to estimate the status of the next timestamp $x_{t+1}$. In practice, these two tasks are expected to work jointly to undertake automatic performance monitoring on the KPIs. Most of the KPIs are the  reflections of the user behaviors, habits, and schedule~\cite{Xu2018}. Since these events are largely repeated periodically, the KPI sequences are mostly stationary and periodic on a daily or weekly basis. Therefore they are believed predictable though the latent factors that impact the statuses are hard to be completely revealed. Four sample sequences are shown in Fig.~\ref{fig:1}. As shown in the figure, the series are mixed with anomalies which are in a rare occurrence and demonstrate drifting patterns.
	
	Performing anomaly detection and prediction on these time series are non-trivial. Firstly, due to the painstaking as well as error-prone annotation, it is unrealistic to expect a large number of labeled data available to train a detection model. As a result, unsupervised anomaly detection is preferred. Secondly, since the anomalies appear in various forms, the prediction model is expected to be robust to the noise. In the existing solutions, these two issues are addressed separately. In unsupervised anomaly detection, generative models such as variational auto-encoders (VAEs)~\cite{Kingma2013} are employed~\cite{Fan2020}. The time series are sliced by a sliding window~\cite{Chen2019,Xu2018}. The sequence within one window is therefore encoded/decoded by VAE. Anomaly statuses are identified as they are considerably different from the decoded normal statuses. Since all the statuses including anomalies in the sequence are fed to train the model, the interference from the noise and anomalies becomes inevitable. This leads to unstable performance. RNN based models are usually adopted in prediction. However, due to the high model complexity of LSTM~\cite{Hochreiter1997}, they are sensitive to the anomalies and noise. This problem is alleviated by ensemble learning~\cite{Xue2015,Zameer2017}, whereas several folds of computational overhead are induced.
	
	Different from the existing solutions, a joint model called predictor and anomaly detector (PAD) is proposed in this paper. In our solution, VAE and LSTM are integrated as a whole to address both robust prediction and unsupervised anomaly detection. In addition, spectral residual (SR)~\cite{cvpr07:hou} is plugged into the processing pipeline to boost the performance. Specifically, a weight has been assigned by SR to the status at each time slot to indicate the degree of being a normal status.
	
	The advantages of such a framework are at least two folds.
	\begin{itemize}
		\item Firstly, VAE and LSTM have been integrated seamlessly to fulfill both anomaly detection and prediction. The VAE block is in charge of anomaly detection and LSTM is adopted for prediction. On the one hand, VAE considerably reduces the impact from the anomalies and noise on the prediction block. On the other hand, LSTM helps VAE to maintain the long term sequential patterns that are out of the VAE encoding window. This design leads to the better performance for both tasks than they work alone.
		\item Secondly, spectral residual analysis is adopted as a pre-processing step in the whole pipeline. It helps to suppress the apparent anomalies and therefore alleviates their interference to the training of VAE and LSTM.
	\end{itemize}
	
	Although our model is conceptually similar as the models used in anomaly detection~\cite{Zhang2019} and natural language processing~\cite{Bowman2016}, which integrate both recurrent neural network (RNN) and VAE~\cite{Chung2019}, the structure of our model is different. Moreover, our model is capable of robust prediction. To the best of our knowledge, this is the first work that addresses both prediction and unsupervised anomaly detection under one joint model.
	
	The remainder of this paper is organized as follows. Related works about prediction and unsupervised anomaly detection are presented in Section~\ref{sec:rela}. The proposed model, namely PAD is presented in Section~\ref{sec:method}. The effectiveness of our approach both for prediction and anomaly detection is studied on two datasets in Section~\ref{sec:exp}. Finally, Section~\ref{sec:conc} concludes the paper.
	
	\section{Related Work}
	\label{sec:rela}
	\subsection{Prediction on Time Series}
	Prediction on the time series is an old topic as well as a new subject. On the one hand, it is an old topic in the sense it could be traced back to nearly one century ago~\cite{DeGooijer2006}. In such a long period of time, classic approaches such as ARIMA~\cite{Box1976}, Kalman Filter~\cite{Harvey1990}, and Holt-Winters~\cite{Kalekar2004} were proposed one after another. The implementations of these classic algorithms are found from recent packages such as Prophet~\cite{Facebook2017} and Hawkular~\cite{RedHat2014}. Although efficient, the underlying patterns are usually under-fit due to the low model complexity. On the other hand, this is a new issue in the sense that the steady growth of the big volume of IT operation data, which are mixed with the noise and anomalies, impose new challenges to this century-old issue.
	
	Recently LSTMs~\cite{Hochreiter1997} are adopted for prediction due to their superior capability in capturing long-term patterns on temporal data. A recent work integrates attention mechanism into RNN as the nonlinear autoregressive exogenous model~\cite{Qin2017}. However, RNN turns out to be sensitive to the anomalies and noise. In order to enhance its robustness, the constraint on the excessive inputs or gradients is introduced during the training~\cite{Connor1994,Lee2009}. However, only limited improvement is observed. Moreover, in~\cite{Xue2015,Zameer2017}, multiple prediction models are trained from one time series, and the prediction is made by integrating multiple predictions into one. LSTM is also modified to perform online prediction in~\cite{Guo2016}. The learned model is adapted to the emerging patterns of time series by balancing the weights between the come-in gradient and historical gradients.
	
	\subsection{Anomaly Detection on Time Series}
	First of all, data annotation on the time series is expensive and error-prone. It also requires the annotator to be familiar with a specific domain~\cite{Ahmad2017,Canizo2019}. In addition, compared to the normal statuses, the anomalies are in rare occurrences. This makes the training suffer from the class imbalance when it is addressed as a classification problem. Therefore, most of the research in the literature address the detection in an unsupervised manner.
	
	The first category of anomaly detection approaches is built upon the prediction. Specifically, when the status is far apart from the predicted status at one timestamp, it is considered an anomaly. In~\cite{Yaacob2010}, ARIMA is employed for prediction. The detection is fulfilled based on the predicted status. However, due to the poor prediction performance of ARIMA, precise anomaly detection is not achievable. Recently, a stacked LSTM~\cite{Malhotra2015} is proposed to perform anomaly detection due to its good capability in capturing patterns from the time series with lags of unknown duration. However, the uncertainty of the prediction model itself is overlooked in this approach. To address this issue, the research from Uber introduced Bayesian networks into LSTM auto-encoder. MC dropout is adopted to estimate the prediction uncertainty of the LSTM auto-encoder~\cite{Laptev2017,Zhu2017}. In addition to the uncertainty of the prediction model, historical prediction errors are considered in a recent approach from NASA~\cite{Hundman2018}. In recent approach~\cite{Lin2020}, LSTM is placed in-between the encoder and the decoder of VAE to predict the next embedding. As LSTM is sensitive to the noise and anomalies, its detection performance is unstable. All the aforementioned detectors rely largely on the performance of the prediction. Inferior performance is observed when the time series show drifting patterns or they are mixed with the noise.
	
	Another category of anomaly detection approaches is built upon the assumption that the distribution of anomalies considerably differs from the normal status and they are in a rare occurrence. In the early approaches of this category, the detection is simply based on 3$\sigma$ rule. Under the similar principle, recent approaches SPOT and DSPOT~\cite{DeHaan2006} are proposed based on Extreme Value Theory. Since the normal statuses may be under several different distributions, Gaussian mixture model (GMM) is proposed to adapt to the data distribution. Different from GMM, one-class SVM~\cite{Campbell2001} transforms the sequential data directly into another space by kernel function, where the anomalies could be discriminated by the classification plane. However, the performance of one-class SVM turns out to be poor mainly because the kernel is fixed. Due to its superior capability in non-linear mapping, VAE is introduced in~\cite{Chen2019,Xu2018} to encode the sequence within each sliding window. LSTM is incorporated as the layers of VAE~\cite{ZhangLi2019,Niu2020,Park2018} to capture the long term patterns in a series. Since LSTM is sensitive to noise, the model is unstable and hard to converge during the training stage. Observing that the anomalies could be distinguished at the early steps of a decision tree, iForest~\cite{Ding2013} and robust random cut forest~\cite{Guha2016} are proposed. Recently, spectral residue analysis (SR), which is originally used on saliency detection in computer vision, is introduced for anomaly detection~\cite{Ren2019}. As the saliency is inconsistent with the anomaly in some cases, it turns out to be fast, whereas the performance is unstable.

	Recently, supervised solution is also seen in the literature~\cite{Ren2019}. In order to fulfill the training, the anomalies are synthesized and injected into the original time series in SR-CNN~\cite{Ren2019}. Additionally, a large amount of extra time series are synthesized and used in the training to prevent the model from overfitting. However, the synthesized statuses cannot reflect the real distribution of anomalies in practice, which leads to unstable performance of the trained model.
	
	In the above VAE based detection approaches, the judgment is made mainly based on the distribution difference between the normal and abnormal statuses. Since the noise and anomalies are also fed into the model training, these signals are unexpectedly reconstructed as the normal ones. As a result, the boundary between normal and abnormal statuses is blurred. To address this issue, SR~\cite{cvpr07:hou} is integrated into our model to suppress the anomalies before they are fed into the VAE block.
	
	\section{The Proposed Model}
	\label{sec:method}
	In this section, a framework integrated with LSTM and VAE for both unsupervised anomaly detection and robust prediction is presented.
	
	\subsection{Preprocessing}
	\label{sec:pre}
\begin{figure}[t]
		\centerline{\includegraphics[width=0.4\textwidth]{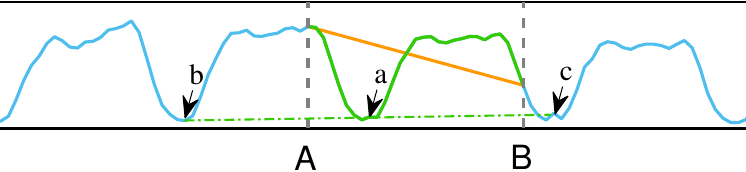}}
		\caption{Fill missing statuses by our way. The statuses between $A$ and $B$ are missing. The linear interpolation is shown in orange. The linear interpolation with the same time slot from the adjacent periods is shown in green, i.e., fill $a$ with the interpolation of $b$ and $c$.}
		\label{fillmissing}
	\end{figure}

	As shown in Fig.~\ref{fig:1}, it is not unusual that the operation statuses are missing on some timestamps due to the sudden server down or network crashes. The conventional schemes are zero filling and linear interpolation, which could damage the periodicity of time series. In our solution, the missing statuses are filled with adjacent periods. Specifically, when the missing duration is less than or equal to \textit{M} time units, the linear interpolation \cite{Davis1975} is performed by filling the missing statuses with the adjacent statuses. When the missing duration is greater than \textit{M} time units, the linear interpolation is performed with the status of the same time slot from the adjacent periods as shown in Fig.~\ref{fillmissing}. In our implementation, we choose one day as the period, which is safe as IT operations are largely relevant to human daily activities. \textit{M} is set to \textit{3} and \textit{7} for hour-level and minute-level series respectively. The periodicity recovery by this way is crucial to the processing in the later stages, such as spectral residual analysis in the frequency domain.
	
	\begin{figure*}[!t]
		\centerline{\includegraphics[width=0.65\linewidth]{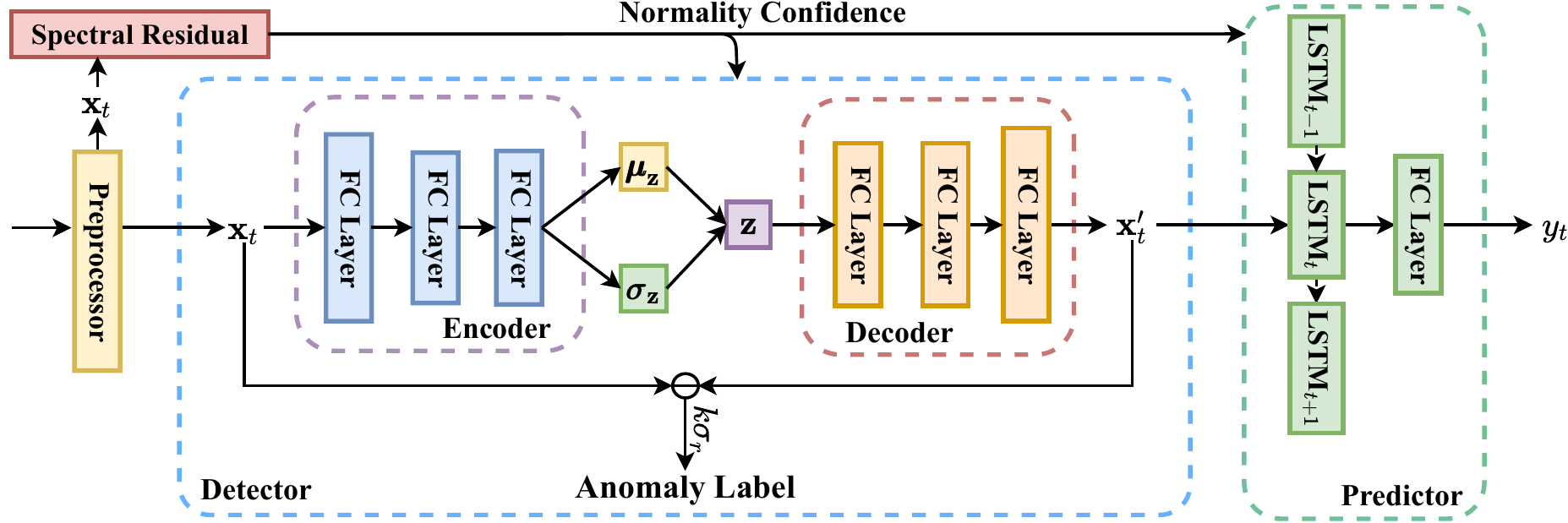}}
		\caption{The structure of our model for anomaly detection and prediction. There are two major blocks in the network, namely the VAE block shown inside the blue dashed box and the LSTM block shown inside the green dashed box. After preprocessing, spectral residual analysis is used to produce the normality confidence for each status in one segment $\textbf x_t$. Segment $\textbf x_t$ is then fed into VAE and LSTM. For anomaly detection, $\textbf x_t$ is reconstructed as $\textbf x_t'$ by VAE. The difference between input status $\textbf x_t$ and $\textbf x_t'$ is transformed into an anomaly label by the threshold $k\sigma_{r}$. $\textbf x_t'$ is also fed into LSTM to estimate $y_t$ for robust prediction.}
		\label{fig:seqvl}
	\end{figure*}
	
	After filling the missing statuses, each time series is undergone z-score normalization. Then each time series is cut into segments with a sliding window. The window size is ${\Omega}$. Similar to the existing works, the step size of the window sliding is fixed to \textit{1}. The statuses in one segment are given as $\textbf x_t=\{x_{t-\Omega+1},\cdots,x_{t}\}$. After segmenting the time series with the sliding window, the time series is decomposed into a collection of segments viz., $G=\{\textbf x_{\Omega},\cdots,\textbf x_t,\cdots,\textbf x_n\}$. Every neighboring \textit{L} segments are organized into a sequence and are fed to the network for training. Such that the temporal clues beyond one segment are maintained.
	
	\subsection{Normality Confidence Weighting}
	Spectral residual (SR) analysis is a traditional signal processing tool. It has been shown useful in identifying salient/ irregular patterns in 1D or 2D signals~\cite{cvpr07:hou}. In recent work~\cite{Ren2019}, it has been employed in the anomaly detection on time series data~\cite{Ren2019} for its efficiency. Similar as~\cite{Ren2019}, SR is employed in our processing pipeline. However, different from~\cite{Ren2019}, it is only employed to assign a normality weight to the status of each timestamp. This weight will be later used to assist the training of our detection and prediction blocks.
	
	Firstly, the log amplitude spectrum of a segment $\textbf{x}_t$ is obtained by \textit{Fourier} transform and log transformation. In the second step, the spectral residual is obtained by subtracting the log amplitude spectrum from its mean. Finally, the spectral residual is transformed back to the time domain. It is the resulting 1D saliency map $S(\textbf{x}_t)$.
	Given the last point $S(x_t)$ and the local average of the last point $\overline{S(x_t)}$ in the saliency map $S(\textbf{x}_t)$, the normality confidence at timestamp $t$ is estimated as
	
	\begin{equation}
		w_n(x_t) = 1-\frac{1}{1+\exp(-(D(x_t)-D_0))}, 	\label{eqn:nconfidence}
	\end{equation}
	\begin{equation}
		\mbox{where } D(x_t) = \frac{S(x_t)-\overline{S(x_t)}}{\overline{S(x_t)}}. \nonumber
	\end{equation}
	
	In Eqn.~\ref{eqn:nconfidence}, $D(x_t)$ indicates the degree that the status at $t$ differs from normal and $D_0$ is a constant. As a result, $w_n(x_t)$ basically indicates the confidence that $x_t$ is normal. Its value ranges from \textit{0} to \textit{1}. $w_n(x_t)$ of the anomalies is expected to be close to \textit{0}. As we show later, this confidence score will be integrated into the VAE-LSTM learning framework to alleviate the interference from the anomalies. In addition, the confidence scores of statuses in segment $\textbf x_t$ are given as $\textbf{w}_n$.
	
	\subsection{Anomaly Detection}
	In this paper, we aim to address anomaly detection and prediction under one framework. Let's consider anomaly detection first. As shown in Fig.~\ref{fig:1}, most of the anomalies appear as isolated points. Assuming that 1. the latent variable of the segment $\textbf x_t$, namely $\textbf z$ follows multivariate standard Gaussian distribution $p_\theta (\textbf z)=\mathcal N(\textbf 0, \textbf I)$ and 2. the anomalies are in a rare occurrence. The status $\textbf x_t$ can be largely reconstructed by
	\begin{equation}
		\textbf x_t' = \text{VAE}(\textbf x_t).
		\label{eqn:vae}
	\end{equation}
	
	In VAE, as an approximation to the intractable true posterior distribution $p_\theta(\textbf z|\textbf x)$, the approximate posterior distribution is assumed to follow a diagonal Gaussian distribution $q_\phi(\textbf z|\textbf x)=\mathcal N(\pmb \mu_{\textbf z},\pmb \sigma_{\textbf z}^2\textbf I)$, which is fitted by the encoder. Therefore, on the encoder side, one segment $\textbf x_t$ is encoded into $\pmb \mu_{\textbf z}$ and $\pmb \sigma_{\textbf z}$ by a three-layer encoder. On the decoder side, sampled from $\mathcal N(\pmb \mu_{\textbf z},\pmb \sigma_{\textbf z}^2\textbf I)$, $\textbf z$ is decoded into $\textbf x'_t$ with a symmetric structure as the encoder. Based on the evidence lower bound as~\cite{Kingma2013,An2015}, our VAE is trained with loss function shown in Eqn.~\ref{eqn:lossvae}.
	\begin{equation}
		\begin{aligned}
			\mathcal L_{\text{VAE}}(\textbf x_t)&=\|\textbf{w}_n\!\circ\!(\textbf x_t-\textbf x'_t)\|_2^2 + \beta\,\overline{\textbf{w}_n}\,\text{KL}\Big(\mathcal{N}(\pmb \mu_{\textbf z},\pmb \sigma_{\textbf z}^2\textbf I)\big\Vert \mathcal N(\textbf 0, \textbf I)\Big)\\
			&=\|\textbf{w}_n\!\circ\!(\textbf x_t-\textbf x'_t)\|_2^2 + \frac{\beta\,\overline{\textbf{w}_n}}{2}\Big(-\log \pmb \sigma_{\textbf z}^2 + \pmb \mu_{\textbf z}^2 + \pmb \sigma_{\textbf z}^2 - 1\Big),
		\end{aligned}
		\label{eqn:lossvae}
	\end{equation}
	where the first term is the reconstruction loss of $\textbf x_t$ and the second term is Kullback-Leibler (KL) divergence between $q_\phi(\textbf z|\textbf x)$ and $p_\theta(\textbf z)$. $\textbf{w}_n$ is the normality confidence of statuses in segment $\textbf{x}_t$ (Eqn.~\ref{eqn:nconfidence}) and $\overline{\textbf{w}_n}$ is the average over $\textbf{w}_n$. $\beta$ is a hyper-parameter to balance the reconstruction accuracy and the consistency between the learned and the assumed data distribution~\cite{Higgins2017}. The integration of normality confidence tunes down the impact from anomalies during the training as they hold lower weight in Eqn.~\ref{eqn:lossvae}. The first term in Eqn.~\ref{eqn:lossvae} regularizes how well the VAE fits the training data. While the second term in the equation emphasizes the generalization of VAE over latent distribution. $\beta$ in Eqn.~\ref{eqn:lossvae} is a hyper-parameter to balance these two competing loss functions.
	
	Since the anomalies are in a rare occurrence, the distribution of these statuses is different from those of the normal statuses. In the ideal case, these anomalies are not recovered by the decoder. Following the practice in \cite{Xu2018}, the anomaly detection is made based on the difference between the last status in the segment $\textbf x_t$ and the recovered segment $\textbf x_t'$. Namely the difference between $x_t$ in $\{x_{t-\Omega+1},\cdots,x_t\}$ and $x_t'$ in $\{x_{t-\Omega+1}',\cdots,x_t'\}$ is checked. However, different from \cite{Xu2018}, $x_t$ is viewed as abnormal when the absolute error of $x_t$ from $x'_t$ is higher than $k\sigma_{r}$. $k$ is fixed on all the time series from one evaluation dataset and $\sigma_{r}$ is the standard deviation of the absolute errors of $x_t$ from $x'_t$. Compared to \cite{Xu2018}, such a threshold scheme adapts well to the different distributions of the absolute errors. The detection is shown as the middle block in Fig.~\ref{fig:seqvl}.
	
	Due to the symmetric structure of VAE, the size of the input layer of encoder and the output layer of decoder are set to be the same as window size $\Omega$. ReLU is selected as the activation function for both layers. The number of $\textbf z$ dimensions is set to $K$. The layer of $\pmb \mu_{\textbf z}$ and the layer of $\pmb \sigma_{\textbf z}$ which learns $\log \pmb \sigma_{\textbf z}$ to cancel the activation function, are both fully-connected layers. Because of the symmetry of the auto-encoder, the hidden layers of the encoder and the decoder are both two layers with the ReLU activation function, each of which is with $h_l$ units. 
	
	To this end, the status at timestamp $t$ could be reconstructed by VAE. Since only the normal statuses are encoded and decoded, anomaly detection is as easy as checking the difference between the decoded status and the input status. However, VAE alone is unable to fulfill the prediction since VAE is unable to encode/decode a future status $x_{t+1}$ outside the window. 
	Meanwhile, the recovered $\textbf x'_t$ is expected free of anomalies. If $\textbf x'_t$ is used for prediction, the prediction block becomes insensitive to the noise and possible anomalies. In the following, we are going to show how the output from VAE is capitalized for prediction by LSTM.
	
	\subsection{Prediction}
	LSTM is employed in our design to fulfill prediction. As shown in the right part of Fig.~\ref{fig:seqvl}, LSTM takes the output from the VAE block, and it is expected to predict $x_{t+1}$ based on $\textbf x'_t$. Namely, the loss function is given as
	\begin{equation}
		\mathcal {L}_{\text{LSTM}}(\textbf x_t', x_{t+1})=\overline{\textbf{w}_n}\|x_{t+1}-y_t\|_2^2,
		\label{eqn:losslstmn}
	\end{equation}
	where $y_t$ is the predicted status from LSTM. The loss function simply measures the mean squared error between the true status at timestamp $t+1$ and the predicted status $y_t$. In Eqn.~\ref{eqn:losslstmn}, the normality confidence derived in Eqn.~\ref{eqn:nconfidence} is also integrated. Namely, the average confidence $\overline{\textbf{w}_n}$ for segment $\textbf{x}_t$ is used to weigh the loss function. The contribution of anomalies to the loss function is therefore tuned down.
	
	The LSTM block takes $\textbf x_t'$ as input and fulfills the prediction. Specifically, given the output and the state of the previous timestamp are $\textbf h_{t-1}$ and $\textbf c_{t-1}$, the output and the state of the current timestamp from LSTM are $\textbf h_{t}$ and $\textbf c_{t}$ respectively~\cite{Hochreiter1997}. Namely we have
		\begin{equation}
			\textbf h_{t},\textbf c_{t} = \text{LSTM}(\textbf x_t', \textbf h_{t-1}, \textbf c_{t-1}).
	\end{equation}
	
	In order to map $\textbf h_t$ to the predicted status $y_t$, a fully connected layer is attached to the LSTM block. The predicted status $y_t$ is computed with Eqn.~\ref{eqn:lstm7}.
	\begin{equation}
		y_t = \textbf w_y\textbf h_t+b_y
		\label{eqn:lstm7}
	\end{equation}
	
	During the training of the whole network, the loss functions of prediction and unsupervised anomaly detection should be balanced. So the overall loss function for predictor and anomaly detector (PAD) is
	\begin{equation}
		\begin{aligned}
			\mathcal {L}_{\text{PAD}}(\textbf x_t, x_{t+1})&=\mathcal {L}_{\text{VAE}}(\textbf x_t) + \lambda \mathcal {L}_{\text{LSTM}}(\textbf x_t', x_{t+1})\\
			&=\|\textbf{w}_n\!\circ\!(\textbf x_t-\textbf x'_t)\|_2^2 + \beta\,\overline{\textbf{w}_n}\,\text{KL}\Big(\mathcal{N}(\pmb \mu_{\textbf z},\pmb \sigma_{\textbf z}^2\textbf I)\Big\Vert \mathcal N(\textbf 0, \textbf I)\Big)
			\\
			&\,\,\,\,\,\,+\lambda\,\overline{\textbf{w}_n}\|x_{t+1}-y_t\|_2^2,
			\label{eqn:lossSeqVL}
		\end{aligned}
	\end{equation}
	where $\mathcal {L}_{\text{VAE}}(\textbf x_t)$ is the loss function of unsupervised anomaly detection and $\mathcal {L}_{\text{LSTM}}(\textbf x_t', x_{t+1})$ is the loss function of robust prediction. $\lambda$ is a hyper-parameter to balance the training over these two tasks. The KL divergence for the VAE block (given in Eqn.~\ref{eqn:lossvae}) emphasizes the latent constraint of VAE, which basically indicates how robust the reconstructed output $\textbf x_t'$ from VAE when anomalies or noise appear. When its weight in the overall loss function ($\mathcal {L}_{\text{PAD}}$) is high, the output ($\textbf x_t'$) from the VAE block is cleaner. The cleaned statuses are input to LSTM for prediction. Therefore, as the key hyper-parameter of our model, $\beta$ controls the generalization of anomaly detection block and impacts the robustness of prediction as well.
	
	Although training is required in our network, it is essentially an unsupervised approach in the sense that no anomaly annotation is required. The training largely allows the VAE and LSTM to adapt to the distribution of the time series. The prediction block can be viewed as a natural extension over the anomaly detection block as we make full use of the output from VAE. The reconstructed segment from the VAE reduces the noise in the raw input data. With the clean input, the LSTM block is able to capture the regular temporal patterns. On the one hand, it fulfills the prediction task with less interference from noise. On the other hand, it propagates the long term temporal clues back to the VAE block to boost its performance in anomaly detection. As will be revealed in the experiments, both blocks perform better than they work alone. 
	
	To the best of our knowledge, this is the first piece of work that integrates prediction and anomaly detection into one framework. The anomaly detection block and the prediction block are built upon each other. This is essentially different from \cite{Laptev2017,Zhu2017,Hundman2018}, which are trained solely for prediction but used for detection. Our model is essentially different from VAE-LSTM in~\cite{Lin2020} in the sense that LSTM is not employed to perform anomaly detection. Instead, it is in charge of prediction only. 
	
	\section{Experiments}
	\label{sec:exp}
	\begin{table*}[!t]
		\caption{Summary over the datasets}
		\label{tab:data}
		\begin{center}
			\begin{tabular}{|c|c|c|c|c|} \hline
				\textbf{Dataset}  &\textbf{{\#} Series} &\textbf{{\#} Time-stamps} &\textbf{{\#} Anomalies} &\textbf{Granularity}\\
				\hline \hline
				\textbf{KPI}   & 29    & 5,922,913 & 134,114 (2.26\%) & Minute \\
				\textbf{Yahoo} & 367   & 572,966 & 3,896 (0.68\%) & Hour \\ \hline
			\end{tabular}
		\end{center}
	\end{table*}

	\begin{figure*}[!t]
		\centering
		\subfigure[]{\includegraphics[width=0.4\linewidth]{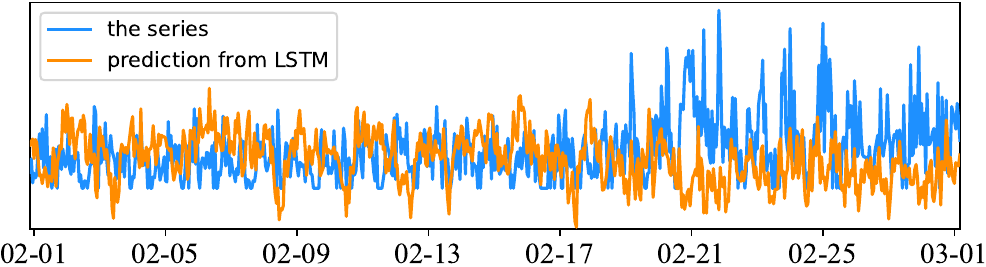}}
		\hspace{0.4cm}
		\subfigure[]{\includegraphics[width=0.405\linewidth]{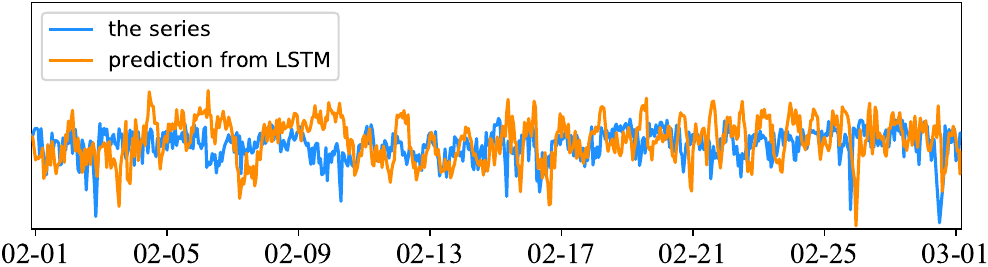}}
		\\
		\hspace{-0.2cm}
		\subfigure[]{\includegraphics[width=0.4\linewidth]{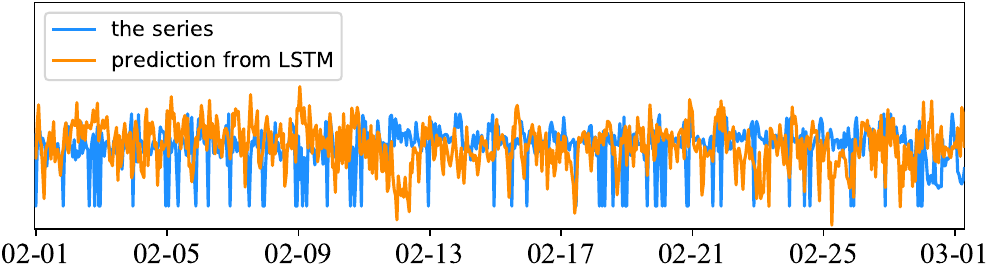}}
		\hspace{0.7cm}
		\subfigure[]{\includegraphics[width=0.38\linewidth]{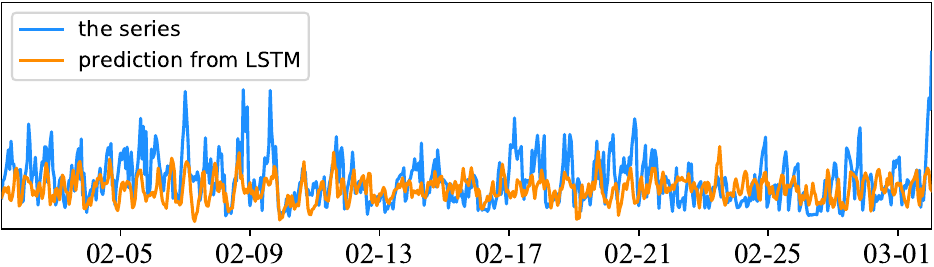}}
		
		\centering
		\caption{Four sample series (in blue color) from dataset \textbf{Yahoo} along with their prediction results (in orange color) obtained by LSTM. As shown in the figures, the time series demonstrate irregular patterns which are hardly predictable. The prediction results from LSTM turn out to be very poor.}
		\label{fig:yahoo_pred}
	\end{figure*}
	
	In this section, the effectiveness of the proposed approach is studied in comparison to the approaches that are designed for prediction and  unsupervised anomaly detection in the literature. Datasets \textbf{KPI}~\cite{AIOpsChallenge2017} and \textbf{Yahoo}~\cite{YahooLabs2015} are adopted in the evaluation. \textbf{KPI} dataset was released by the AIOps Challenge Competition~\cite{AIOpsChallenge2017}. It consists of desensitized time series of KPIs from real-world applications and services. The raw data were harvested from Internet companies such as Sogou, eBay, and Alibaba. They are all minute-level operations time series. In our evaluation, this dataset is used for both anomaly detection and prediction. \textbf{Yahoo} dataset is released by Yahoo Labs for anomaly detection evaluation. It consists of both real and synthetic time series. Most of the time series from this dataset demonstrate irregular patterns across the whole time span. Such kind of patterns is believed unpredictable~\cite{Ren2019}. Four sample series from $\textbf{Yahoo}$ along with the prediction results from LSTM are shown in Fig.~\ref{fig:yahoo_pred}. As a result, this dataset is used to evaluate anomaly detection performance only following the convention in the literature~\cite{Ren2019}. The brief information about these two datasets is summarized in Tab.~\ref{tab:data}.
	
	On \textbf{KPI} dataset, $D_0$ in Eqn.~\ref{eqn:nconfidence} is empirically set to \textit{4.1}. $\beta$ in Eqn.~\ref{eqn:lossvae} is set to \textit{0.01}. As $\beta$ is crucial to our model, this setting will be validated by an ablation study. $\lambda$ in Eqn.~\ref{eqn:lossSeqVL} is set to \textit{1}. The segment sequence length $L$ is set to \textit{256}. Other hyper-parameters are configured according to~\cite{Xu2018}. On \textbf{Yahoo} dataset, $D_0$ in Eqn.~\ref{eqn:nconfidence} is set to \textit{3.1}. $\lambda$ is set to \textit{10}. The window size $\Omega$ is set to \textit{30}. $h_l$ and the size of $\textbf h_t$ are set to \textit{24}. The learning rate is set to \textit{1e-2}. The rest of the configurations on the training is kept the same as on \textbf{KPI} dataset. The configurations in SR is set following \cite{Ren2019}. For the readers who want to repeat our work, the following principles in parameter-tuning are recommended.
	\begin{enumerate}
		\item Hyper-parameter $\lambda$ is set according to the preference of the two tasks, namely prediction and anomaly detection. In our practice, anomaly detection is preferred over prediction;
		\item Hyper-parameter $\beta$ is set as a trade-off between accuracy and generalization of PAD;
		\item $D_0$ is set according to the performance of SR on the dataset.
	\end{enumerate}

	\subsection{Evaluation Protocol}
	Following~\cite{Ren2019,AIOpsChallenge2017}, the first half of the time series is used to train the model, while the second half is used for evaluation in both tasks. In the prediction task, Mean Squared Error (MSE) (Eqn.~\ref{eqn:mse}), Root Mean Squared Error (RMSE) (Eqn.~\ref{eqn:rmse}) and Mean Absolute Error (MAE) (Eqn.~\ref{eqn:mae}) are used in the evaluation. 
	\begin{equation}
		\text{MSE} = \frac{1}{n-\Omega}\sum_{t=\Omega}^{n-1}(x_{t+1}-y_t)^2
		\label{eqn:mse}
	\end{equation}
	\begin{equation}
		\text{RMSE} = \sqrt{\frac{1}{n-\Omega}\sum_{t=\Omega}^{n-1}(x_{t+1}-y_t)^2}
		\label{eqn:rmse}
	\end{equation}
	\begin{equation}
		\text{MAE} = \frac{1}{n-\Omega}\sum_{t=\Omega}^{n-1}|x_{t+1}-y_t|,
		\label{eqn:mae}
	\end{equation}
	where $\Omega$ is the window size and $y_t$ is the predicted value from the LSTM block. The final performance score is obtained by taking the average over MSE, RMSE, and MAE respectively of all the time series from one dataset.
	
	In addition, as the anomalies in one time series are believed unpredictable, the scores of MSE, RMSE, and MAE are reported under three setups. Specifically, in Setup-1, the performance of our prediction approach is studied when the anomalies are considered as part of the ground-truth. In Setup-2, the time slots in the ground-truth where anomalies appear are replaced with estimated normal status with the scheme presented in Section~\ref{sec:pre}. In Setup-3, the time slots where anomalies appear are simply ignored in the evaluation.
	
\begin{figure}
		\centering
		\includegraphics[width=0.8\linewidth]{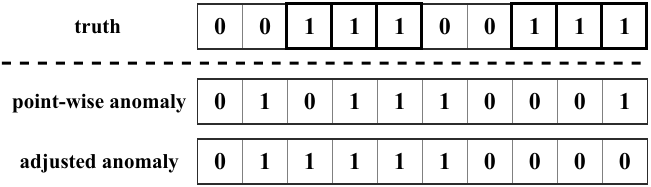}
		\centering
		\caption{The illustration of the strategy used in the evaluation of anomaly detection.}
		\label{fig:eval}
	\end{figure}

	For the anomaly detection evaluation, the operators in practice do not care about whether an anomaly is detected successfully at the moment it appears. Instead they care about in which time span an anomaly is successfully detected within a small tolerable delay. As a result, the strategy in~\cite{Ren2019,Xu2018,AIOpsChallenge2017} is adopted in the evaluation. As illustrated in Fig.~\ref{fig:eval}, if the model detects anomalies no later than the delay after the start timestamp of the anomaly interval, each timestamp in the anomaly interval is viewed as a true positive. Otherwise, each timestamp in the anomaly interval is counted as a false negative. The delay for adjustment is set to \textit{3} and \textit{7} for hour-level and minute-level datasets respectively. We evaluate the detection performance with precision, recall and F$_1$-score by Eqn.~\ref{eqn:p}, Eqn.~\ref{eqn:r}, and Eqn.~\ref{eqn:f1}.
	
	\begin{equation}
		\text{precision} = \frac{\text{\#True positive}}{\text{\#True positive}+\text{\#False positive}}
		\label{eqn:p}
	\end{equation}
	\begin{equation}
		\text{recall} = \frac{\text{\#True positive}}{\text{\#True positive}+\text{\#False negative}}
		\label{eqn:r}
	\end{equation}
	\begin{equation}
		\text F_1\text{-score} = \frac{2\times\text{precision}\times\text{recall}}{\text{precision}+\text{recall}}
		\label{eqn:f1}
	\end{equation}
	
	\subsection{Ablation Study}
	Hyper-parameter $\beta$ in Eqn.~\ref{eqn:lossvae} balances the reconstruction accuracy and latent constraint of VAE. It plays a crucial role in our model. In the first experiment, we study the performance of our model as this parameter setting varies from \textit{0.01} to \textit{10.0}. In this study, only the performance on prediction is reported. The similar performance trend is observable for the anomaly detection task. The results are presented in Tab.~\ref{tab:beta_predict}.
	
	As shown in the table, the performance of prediction is poor when $\beta$ is set to large values. Large $\beta$ tunes down the importance of prediction accuracy in the loss function. Moreover, large $\beta$ emphasizes the consistency between learned data distribution with the assumed data distribution, which leads to the extremely blurry reconstruction. The same phenomenon is observed in anomaly detection. As a consequence, $\beta$ in Eqn.~\ref{eqn:lossSeqVL} is fixed to \textit{0.01} on both \textbf{KPI} and \textbf{Yahoo} datasets.
	
	\begin{table}[!b]
		\centering
		\caption{The prediction performance of PAD with different $\beta$ on \textbf{KPI} dataset under Setup-2}
		\renewcommand\tabcolsep{3.3pt}
		\begin{tabular}{|l|c|c|c|c|}
			\hline
			\textbf{$\beta$} & \textbf{0.01} & \textbf{0.1} & \textbf{1.0} & \textbf{10.0} \\
			\hline
			\hline
			\textbf{MSE}   & \textbf{0.1086} & 0.1125 & 0.1240 & 0.1627 \\
			\hline
			\textbf{RMSE}  & \textbf{0.2724} & 0.2806 & 0.3007 & 0.3704 \\
			\hline
			\textbf{MAE}   & \textbf{0.1704} & 0.1800 & 0.2007 & 0.2633 \\
			\hline
		\end{tabular}
		\label{tab:beta_predict}
	\end{table}
	\subsection{Robust Prediction}
	\begin{figure}[!t]
		\centering
		\includegraphics[width=0.9\linewidth]{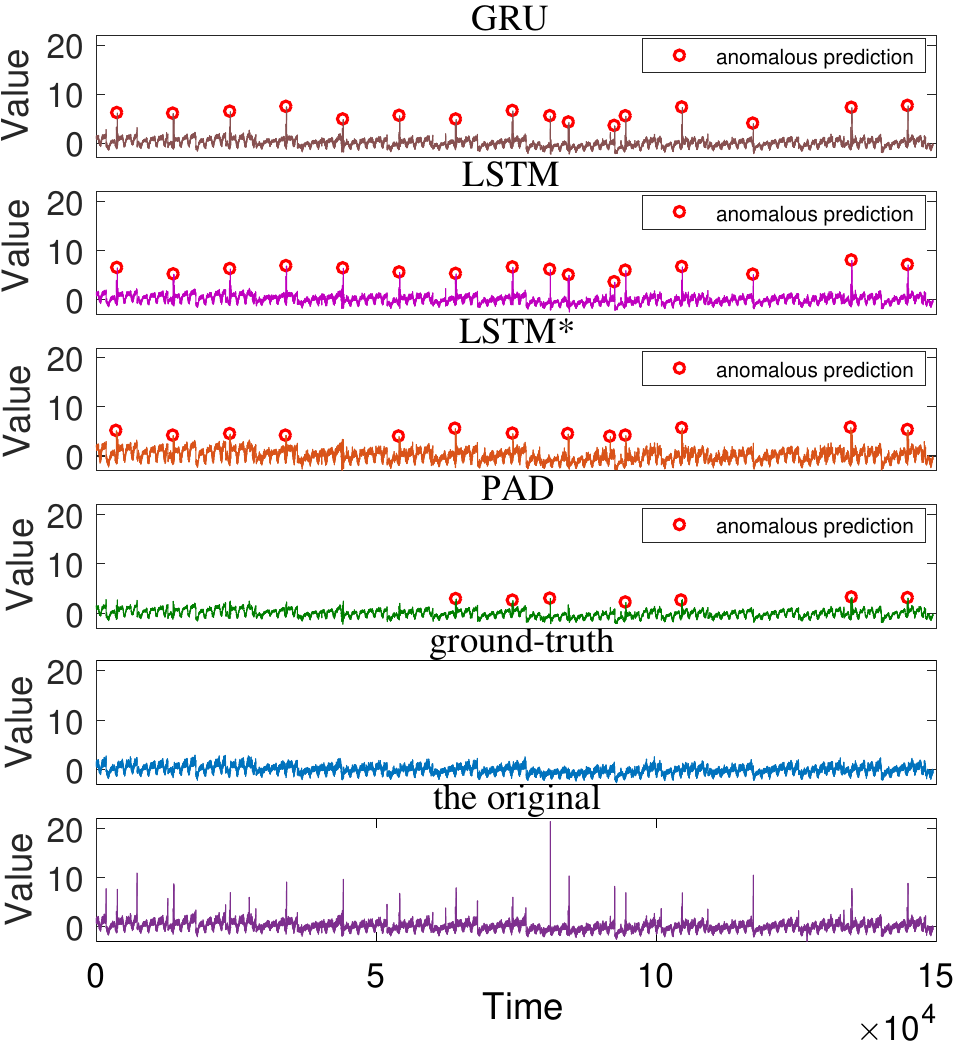}
		\centering
		\caption{The prediction results from GRU, LSTM, LSTM*, and PAD on Sequence-16 from \textbf{KPI} dataset. The original sequences and the ground-truth under Setup-2 are also presented. For GRU, LSTM, and LSTM*, there are many false predictions due to the interference from anomalies during the training.}
		\label{fig:demo}
	\end{figure}
	
	\begin{figure*}[!t]
		\begin{center}
			\subfigure[Measured with MSE~(Eqn.~\ref{eqn:mse})]
			{\includegraphics[width=0.28\linewidth]{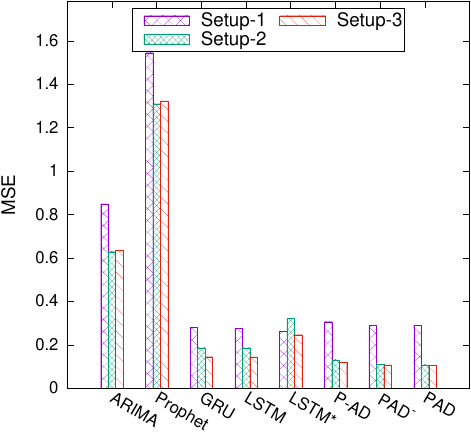}}
			\hspace{0.01in}
			\subfigure[Measured with RMSE~(Eqn.~\ref{eqn:rmse})]
			{\includegraphics[width=0.28\linewidth]{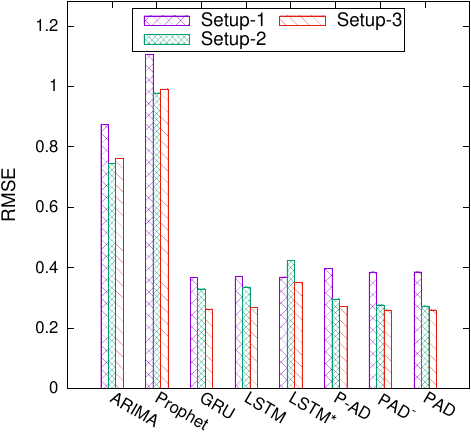}}
			\hspace{0.01in}
			\subfigure[Measured with MAE~(Eqn.~\ref{eqn:mae})]
			{\includegraphics[width=0.28\linewidth]{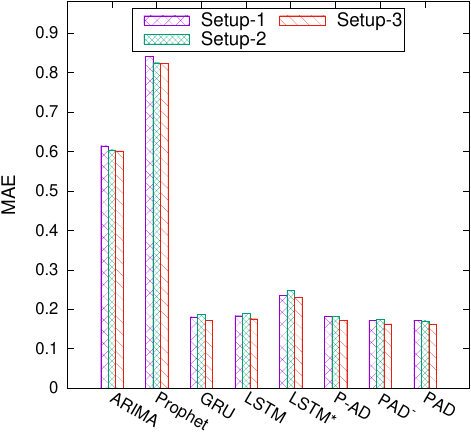}}
		\end{center}
		\caption{The prediction performance of PAD in comparison to ARIMA, Prophet, GRU, LSTM and LSTM* on \textbf{KPI} dataset under three different setups.}
		\label{fig:predict}
	\end{figure*}
	
	\begin{table*}[!t]
		\centering
		\caption{Performance comparison on Anomaly Detection on \textbf{KPI} and \textbf{Yahoo}. The supervised approach is marked with `*'}
		\begin{tabular}{|l|ccc|ccc|}
			\hline
			\multicolumn{1}{|c|}{} & \multicolumn{3}{c|}{\textbf{KPI}} & \multicolumn{3}{c|}{\textbf{Yahoo}} \\
			\hline
			Approach & \multicolumn{1}{c}{F$_1$-score} & \multicolumn{1}{c}{Precision} & \multicolumn{1}{c|}{Recall} & \multicolumn{1}{c}{F$_1$-score} & \multicolumn{1}{c}{Precision} & \multicolumn{1}{c|}{Recall} \\
			\hline \hline
			\textbf{OCSVM}~\cite{Bernhard1999} & 0.183 & 0.144 & 0.251 & 0.026 & 0.013 & 0.803 \\
			\textbf{VAE-LSTM}~\cite{Lin2020} & 0.061 & 0.033 & 0.423 & 0.026 & 0.014 & 0.244 \\
			\textbf{SPOT}~\cite{Siffer2017} & 0.217 & 0.786 & 0.126 & 0.338 & 0.269 & 0.454 \\
			\textbf{DSPOT}~\cite{Siffer2017} & 0.521 & 0.623 & 0.447 & 0.316 & 0.241 & 0.458 \\
			\textbf{DONUT}~\cite{Xu2018} & 0.595 & 0.735 & 0.500 & 0.501 & 0.669 & 0.401 \\
			\textbf{SR}~\cite{Ren2019} & 0.622 & 0.647 & 0.598 & 0.563 & 0.451 & 0.747 \\
			\textbf{VAE}~\cite{Kingma2013} & 0.685 & 0.725 & 0.648 & 0.642 & 0.773 & 0.549 \\
			\textbf{*SR-CNN}~\cite{Ren2019} & \textbf{0.771} & 0.797 & 0.747 & 0.652 & 0.816 & 0.542 \\
			\hline
			\textbf{AD} & 0.726 & 0.884 & 0.615 & 0.737 & 0.806 & 0.678 \\
			\textbf{PAD${}^-$} & 0.711 & 0.757 & 0.670 & 0.734 & 0.881 & 0.630 \\
			\textbf{PAD} & 0.739 & 0.839 & 0.660 & \textbf{0.755} & 0.837 & 0.688 \\
			\hline
		\end{tabular}
		\label{tab:detection}
	\end{table*}
	The prediction performance of our approach is studied on \textbf{KPI} dataset. It is compared to representative approaches in the literature and industry. They are classic approaches such as ARIMA~\cite{Box1976} and Prophet~\cite{Facebook2017}. The latter was recently developed by Facebook. Grid search is adopted in ARIMA within the range of maximum order \textit{5} to fine-tune the hyper-parameters. Prophet runs with default settings. Our approach is also compared to standard Gated Recurrent Unit (GRU)~\cite{Cho2014} and LSTM that is popularly used for prediction. Usually the time series are mixed with the noise and anomalies, which would impact the performance of LSTM. We further show the performance of LSTM with relatively clean data. In this run, LSTM is fed with data of which the apparent anomalies are replaced with estimated normal status values. This run is given as LSTM*. In addition, there are another two runs of our approach are conducted. They are P-AD and PAD${^-}$. P-AD undertakes two-step training. The VAE and LSTM were trained separately. This is to study the effectiveness of joint training. In PAD${^-}$, it was configured with joint training, however without confidence weighting (Eqn.~\ref{eqn:nconfidence}). It is to show the improvement we achieve with the confidence weighting (from SR). For the fair comparison, the same settings for all the hyper-parameters are shared by all the above RNN-based models.
	
	The prediction performance is summarized in Fig.~\ref{fig:predict}. As shown in the figure, the performance from classic approaches is very poor. Both ARIMA and Prophet show high prediction errors. As pointed out in \cite{hyndman2018}, the standard ARIMA normally converges to a constant in long-term prediction when the time series is stationary. In contrast, RNN based models such as LSTM, LSTM*, PAD perform significantly better. In particular, our model PAD achieves the best performance under Setup-2 and Setup-3. Moreover, a wide performance gap is observed between ours and the other RNN based approaches. As the values of anomalies are far from the normal ones, the prediction errors are large under Setup-1 for all the approaches. In this case, the performance gap between our approach and other RNN approaches is narrow. Nevertheless, our approach still shows the best performance in most of the cases.
	
	Compared to LSTM and PAD${}^-$,  PAD shows better performance. This indicates that VAE and SR weighting both help to alleviate the impact from the noise and anomalies to the LSTM block.  Compared to P-AD, PAD also achieves better performance. This shows that the joint training makes the reconstructed output from VAE not only robust, but also beneficial to the prediction. Fig.~\ref{fig:demo} shows a sample sequence from \textbf{KPI} dataset along with the predicted results from GRU, LSTM, LSTM*, and PAD. In general, approaches such as GRU, LSTM and LSTM* are able to predict the status well. However, many anomalies are produced as they are too sensitive to the anomalies and noise in training. In contrast, the sequence predicted by PAD are mixed with considerably fewer anomalies as the interference from anomalies is suppressed by SR and VAE.
	
	\subsection{Unsupervised Anomaly Detection}
	Our anomaly detection approach is compared to the representative approaches in the literature. The considered neural network based approaches include VAE, DONUT~\cite{Xu2018},  SR-CNN~\cite{Ren2019}, and VAE-LSTM~\cite{Lin2020}. The approaches based on conventional models, such as One-Class SVM (OCSVM)~\cite{Bernhard1999}, SR, SPOT and DSPOT~\cite{Siffer2017} are also considered in the comparison. The results of SPOT, DSPOT, SR, and SR-CNN are quoted directly from~\cite{Ren2019}. Among these approaches, VAE shares the same hyper-parameters settings as our VAE block in PAD. Essentially, DONUT is a variant of VAE. Its hyper-parameters on \textbf{KPI} dataset are set according to \cite{Xu2018}. While on \textbf{Yahoo} dataset, the hyper-parameters of DONUT are set to be the same as PAD. As suggested in \cite{Lin2020}, the window size $\Omega$ on $\textbf{KPI}$ and $\textbf{Yahoo}$ is set to \textit{24} and \textit{144} respectively. The rest of the configurations are kept the same as \cite{Lin2020}. For OCSVM, the implementation from \cite{scikit-learn} is adopted. The optimal hyper-parameter $\nu$ is searched in \textit{(0, 1]} to achieve the best F$_1$-score. In SR-CNN, the CNN model used for detection is trained with an extra large amount of time series, in which the anomalies are artificially injected. This is the only supervised approach considered in our study. For our approach, besides the standard configuration, another two runs are pulled out. They are AD and PAD${^-}$. In AD, only VAE block is trained to perform anomaly detection. This is to study the effectiveness of joint training with LSTM. In order to study the performance gain from SR weighting, PAD${^-}$ is pulled out. In this run, SR confidence weighting is turned off.
	
	The performance of anomaly detection from all the aforementioned approaches is presented on Tab.~\ref{tab:detection}. As shown in the table, the proposed approach outperforms all the state-of-the-art unsupervised approaches considerably on both datasets. Its performance remains stable across two different datasets. In contrast, the approaches such as OCSVM, DSPOT, DONUT, and SR demonstrate significant performance fluctuation across different datasets. VAE-LSTM shows very poor performance on both datasets. It turns out to be hard for LSTM to capture the temporal patterns in the embedding space of VAE, since VAE and LSTM are trained separately.
	
	Compared to the run that trains the VAE block alone (AD), PAD shows \textit{2\%} performance improvement on two evaluation datasets. This indicates the LSTM block is able to propagate back the temporal clues to the VAE block to boost its performance. Compared to PAD${}^-$, PAD achieves \textit{2\%} performance improvement due to the SR confidence weighting. Overall, both joint training and SR boost the performance of anomaly detection. Although SR-CNN performs better on \textbf{KPI} dataset, the clean time series are required to support the training. According to~\cite{Ren2019}, \textit{65} million synthesized points are used for training, which are around \textit{10} times bigger than the size of either \textbf{KPI} or \textbf{Yahoo} datasets. In practice, it is unrealistic for each type of time series data to collect a big amount of anomaly free data for training. As a result, our model is more appealing over supervised approaches in practice.
	
	\section{Conclusion}
	\label{sec:conc}
	We have presented our model (PAD) for both robust prediction and unsupervised anomaly detection on IT operations. On the one hand, anomaly detection is fulfilled by VAE with the weighting from spectral residual analysis. On the other hand, the robust prediction is realized by the LSTM block with the reconstructed statuses from VAE. The beauty of this design is that both VAE and LSTM perform better than they work alone for either task. The prediction block (LSTM) takes clean input from reconstructed time series from VAE, which makes it robust to anomalies and noise. Meanwhile, VAE performs better for anomaly detection as LSTM helps to maintain long-term sequential patterns. Normality confidence weighting by spectral analysis further boosts the performance of both. Under this joint learning framework, its performance is close to or even better than the state-of-the-art supervised approach. 
	
	In some scenarios, the KPIs trend may drift gradually. This requires the both prediction and detection models to be updated incrementally. To address this problem, online training is required. In our future work, the online prediction and detection, as well as the theoretical interpretation of our approach are worthwhile to explore.

\section*{Acknowledgment}
We would like to express our sincere thanks to \textit{Bonree Inc}., Beijing, China, for their support of this work. This work is also supported by National Natural Science Foundation of China under grants 61572408 and 61972326, and the grants of Xiamen University 20720180074.

\bibliographystyle{elsarticle-num}
\bibliography{refer.bib}

\par\noindent
\parbox[t]{\linewidth}{
	\noindent\parpic{\includegraphics[height=1.5in,width=1in,clip,keepaspectratio]{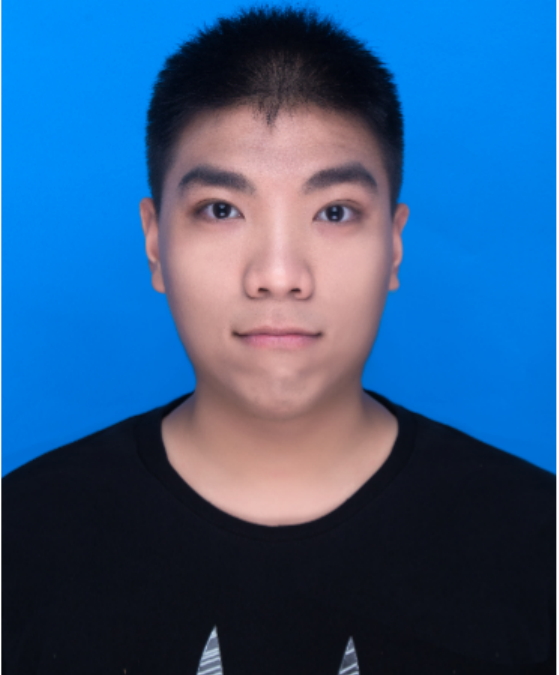}}
	\noindent {\bf Run-Qing Chen}\
	received his Bachelor degree of Information and Computing Science from Nanjing University of Posts and Telecommunications, China in 2018. He is currently a graduate student at Department of Computer Science, Xiamen University. His research interest is anomaly detection in AIOps.}

\vspace{0.3\baselineskip}
\par\noindent
\parbox[t]{\linewidth}{
	\noindent\parpic{\includegraphics[height=1.5in,width=1in,clip,keepaspectratio]{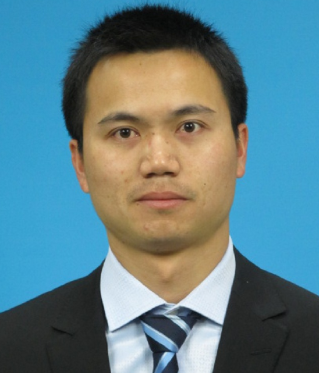}}
	\noindent {\bf Guang-Hui Shi}\
	received his B.Eng. and B.Sci. degrees from Peking University in 2010. He recieved his degree of M.Eng. from Communication and Information System department of Peking University in 2013. He currently works with Bonree Inc., Beijing as a project manager. His team focuses on developing intelligent software tools to monitor the performance of IT systems. Before he joined with Bonree Inc., he worked in Oracle, Beijing as a software engineer.}

\vspace{0.3\baselineskip}
\par\noindent
\parbox[t]{\linewidth}{
	\noindent\parpic{\includegraphics[height=1.5in,width=1in,clip,keepaspectratio]{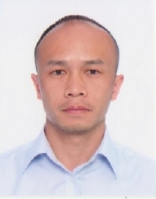}}
	\noindent {\bf Wan-Lei Zhao}\
	received his Ph.D degree from City University of Hong Kong in 2010. He received M.Eng. and B.Eng. degrees in Department of Computer Science and Engineering from Yunnan University in 2006 and 2002 respectively. He currently works with Xiamen University as an associate professor, China. Before joining Xiamen University, he was a Postdoctoral Scholar in INRIA, France. His research interests include multimedia information retrieval and video processing.}

\vspace{0.3\baselineskip}
\par\noindent
\parbox[t]{\linewidth}{
	\noindent\parpic{\includegraphics[height=1.5in,width=1in,clip,keepaspectratio]{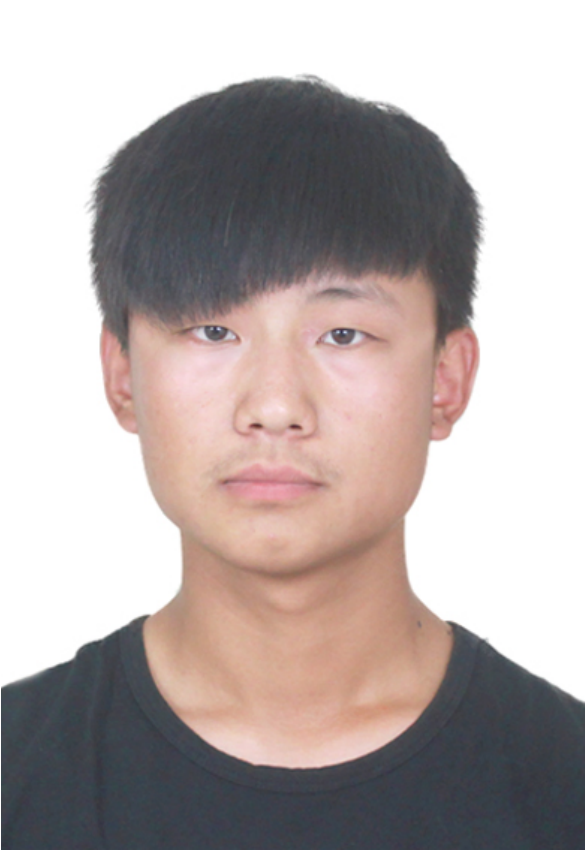}}
	\noindent {\bf Chang-Hui Liang}\
	received his Bachelor degree from Xiamen University, China in 2018. He is currently a graduate student at Department of Computer Science, Xiamen University. His research interest is deep metric learning and fashion search.}

\end{document}